\documentclass[conference]{IEEEtran}
\usepackage{amssymb,amsmath}
\usepackage{epsfig,epstopdf}
\usepackage{subfigure}
\usepackage{multirow}
\usepackage{float}
\usepackage{graphics}
\usepackage[ruled]{algorithm}
\usepackage{algorithmic}
\usepackage{comment}
\usepackage{placeins}
\usepackage{microtype}
\usepackage{multirow}
\usepackage{hyperref}
\usepackage{balance}
\usepackage{color}
\usepackage{graphicx}
\usepackage[normalem]{ulem}

\newcommand{\Rmnum}[1]{\expandafter\@slowromancap\romannumeral #1@}
\newcommand{\tabincell}[2]{
\begin{tabular}{@{}#1@{}}#2\end{tabular}
}

\begin{document}

\title{Cross-modal Zero-shot Hashing}

%\author{Xuanwu Liu$^1$,  Zhao Li$^4$, Jun Wang$^1$, Guoxian Yu$^{1,*}$, Carlotta Domeniconi$^2$, Xiangliang Zhang$^3$\\
%$^1$College of Computer and Information Sciences, Southwest University, China\\
%$^2$Department of Computer Science, George Mason University, USA\\
%$^3$CEMSE, King Abdullah University of Science and Technology, Thuwal, SA\\
%$^4$Alibaba group, Hangzhou, China\\
%Email: \{alxw1007,kingjun,gxyu\}@swu.edu.cn, carlotta@cs.gmu.edu, %zhang@kaust.edu.sa\\}

\author{Xuanwu Liu, Zhao Li, Jun Wang, Guoxian Yu,  Carlotta Domeniconi, Xiangliang Zhang}

% make the title area
\maketitle

\begin{abstract}
Hashing has been widely studied for big data retrieval due to its low storage cost and fast query speed. Zero-shot hashing (ZSH) aims to learn a hashing model that is trained using only samples from seen categories, but can generalize well to samples of unseen categories. ZSH generally uses category attributes to seek a semantic embedding space to transfer  knowledge from seen categories to unseen ones. As a result, it may perform poorly when labeled data are insufficient. ZSH methods are mainly designed for \emph{single-modality} data, which prevents their application to the widely spread multi-modal data.  On the other hand, existing cross-modal hashing solutions assume that all the modalities share the \emph{same} category labels, while in practice the labels of different data modalities may be different. To address these issues, we propose a general  Cross-modal Zero-shot Hashing (CZHash) solution to effectively leverage unlabeled and labeled multi-modality data with different label spaces. CZHash first quantifies the composite similarity between instances using label and feature information. It then defines an objective function to achieve deep feature learning compatible with the composite similarity preserving, category attribute space learning, and hashing coding function learning. CZHash further introduces an alternative optimization procedure to jointly optimize these learning objectives. Experiments on benchmark multi-modal datasets show that CZHash significantly outperforms related
representative  hashing approaches both on effectiveness and adaptability.
%{\color{red}[More strong result statements and improvement can be added.]}
\end{abstract}


\IEEEpeerreviewmaketitle

\section{Introduction}
%With the explosive growth of data by Internet of Things, how to efficiently and accurately retrieve the required information from massive data becomes a hot research topic and has various applications. For example, in information retrieval, approximate nearest neighbor (ANN) search \cite{andoni2006Near} is a fundamental task. Hashing has received increasing attention due to its low storage cost and fast retrieval speed  for  ANN search \cite{Kulis2010Kernelized}.

Hashing aims to compress high-dimensional data into a low-dimensional Hamming space using binary coding, while preserving the proximity among the data  \cite{wang2016L2H,wang2018L2H,Shao2016online}.
Thanks to the compact hash code representation of the original data, hashing can dramatically reduce the storage cost. In addition, hash codes can be leveraged to construct an index and achieve a constant or sub-linear time complexity for data retrieval. For these reasons, hashing has been heavily studied in various domains, ranging from computer vision \cite{Ranjan2015Multi} to bioinformatics \cite{Zhao2019Gene}.

In many domains, the collected data can have different modalities (or feature views). For
example, a web page includes not only a textual description, but also images or videos to supplement its content. These different
types (views) of data are  called \emph{multi-modal} data.
With the rapid growth of multi-modal data, efficient Cross-Modal Hashing (CMH) solutions are in demand. For example, given an image/video of a flower, one may want to retrieve some texts about the flower. How to accomplish cross-modal hashing on multi-modal data is an interesting problem  \cite{Kumar2011Learning,Zhang2014Large,wang2016L2H}.  Depending on whether category labels are used or not, existing cross-modal hashing solutions can be roughly divided into unsupervised, supervised, and semi-supervised (see Table \ref{Table1}).  Unsupervised approaches utilize the underlying data structures, distributions, or topological information to seek
hash functions \cite{Zhou2014Latent,Kumar2011Learning,Ding2014Collective}. Supervised hashing tries to leverage additional information (i.e., semantic labels, ranking between instances)
to improve the performance. Supervised hashing approaches require sufficient labeled instances for model training, while collecting such instances is time-consuming, expensive, and labor intensive (even with crowdsourcing). As such,  semi-supervised hashing solutions  have been proposed to leverage labeled and unlabeled data \cite{Wang2012Semi,liu2019ranking}. However, it is still unfeasible to collect labeled training data for \emph{new concepts} (category labels) in a timely manner. In addition, it is impractical to retrain the hashing functions whenever the retrieval system encounters a new concept.

%For example, Canonical correlation analysis \cite{rasiwasia2010new} maps two modalities, such as visual and textual, into a common space by maximizing the correlation between the projections of the two modalities. Inter-media hashing \cite{Song2013Inter} maps view-specific features onto a common Hamming space by learning linear hash functions with intra-modal and inter-modal consistencies.  %Cross-modal similarity sensitive hashing (CMSSH) \cite{Bronstein2010Data} regards the  hash codes learning as binary classification problems, and efficiently learns the hash functions using a boosting method. Co-regularized hashing \cite{Yi2012Co} learns a group of hash functions for each bit of binary codes in every modal.
%Semantic correlation maximization (SCM) \cite{Zhang2014Large} seeks the hashing functions by maximizing the correlation between two modalities with respect to the semantic labels. Semantics Preserving Hashing (SePH)\cite{Lin2017Cross} generates one unified hash code for all observed views by forcing the semantic consistency between views.

%For example, transferring supervised knowledge (TSK)\cite{Yang2016Zero} takes the semantic vectors {\color{red}What is the 'semantic vectors'?} as a
% bridge to transfer available supervision information from seen
%categories to unseen categories. Binary embedding based Zero-Shot Learning(BZSL)\cite{shen2019scalable} jointly learns two binary coding functions to map class embedding and visual features into the binary embedding space, which could well preserve the discriminate category.
Motivated by the success of zero-shot learning (ZSL) \cite{Lampert2014Attribute,Fu2014Learning,Long2017From}, zero-shot hashing (ZSH) has been recently investigated to deal with  emerging new concepts \cite{Yang2016Zero,shen2019scalable,Zhong2018Attribute}. ZSH aims to encode samples of unseen categories with the hash functions trained using only  samples of seen categories, and by leveraging the techniques of supervised hashing and ZSL. Although ZSH methods \cite{Yang2016Zero,xu2017attribute,guo2017sitnet,lai2018transductive} achieve an impressive performance, they still have some limitations. They only focus on data with one modality, where both the query and the retrieval sets are in the same modality. Attribute-Guided Network for Cross-Modal Zero-Shot Hashing (AgNet) \cite{Zhong2018Attribute} extends ZSH to the multi-modality setting. However, AgNet learns features from image and text modalities separately, and then directly uses the two types of features as input of the hashing quantification network, which may cause incompatibility between  the feature learning  and the quantification process.

When handling multiplicity data (as shown in Table \ref{Table1}), existing ZSH solutions still have the following \textbf{limitations}: (i) They may perform poorly (as we will show in the experiments) when labeled training samples are \emph{insufficient}, since most of them achieve zero-shot learning by embedding semantic labels and rely on sufficient labeled data. (ii) They (including existing CMH solutions) assume that all the modalities share the \emph{same label space}; however,  the labels of different data modalities may not be the same.  For example, an image is tagged with `football', `game' and `grass', but the associated text may be tagged with `football', `FIFA', and `star', where some labels are shared, and the others (`FIFA' and `star') cannot be obtained from the Image modality, and the vice versa.

To address the above limitations, we propose a general  Cross-modal Zero-shot Hashing (CZHash) solution (as illustrated in Fig. \ref{Fig1}) to leverage unlabeled and labeled data annotated with different sets of labels. Specifically, CZHash first quantifies the composite similarity between instances based on the label and feature information of training data from different modalities. Next, it uses the composite similarity to guide feature learning  with a deep neural network.  It then leverages the learned deep features and the class-level attributes to construct the category attribute spaces, which capture
the correlations from seen to unseen classes. CZHash finally introduces a unified object function to jointly account for deep feature learning with composite similarity preserving, category attribute spaces learning, and hashing quantification learning. Our main contributions are summarized as follows:
\begin{enumerate}
\item We study a general cross-modal zero-shot hashing setting (as shown in Table \ref{Table1}) for multi-modality data hashing. Our proposed CZHash unifies compatible deep feature learning, category attribute space learning, and hashing quantification learning in a unified objective.

\item Unlike existing ZSH solutions that embed label information into semantic spaces,  CZHash utilizes label information in different modalities to guide the feature learning process, and further introduces a label and feature information induced composite similarity to leverage labeled and unlabeled data. It then utilizes the learned features with the class-level attributes to construct the category attribute spaces for cross-modal zero-shot hashing.

\item CZHash shows significantly better results than other related and representative cross-modal hashing approaches \cite{Bronstein2010Data,Zhang2014Large,Wu2015Quantized,Lin2017Cross,Jiang2017Deep,liu2019ranking} and zero-shot hashing solution \cite{Zhong2018Attribute} on public multi-modal datasets under different cross-modal hashing scenarios.

\end{enumerate}

%\begin{figure*}[h!tbp]
%\centering
%{\label{Fig1}\includegraphics[width=18cm,height=6cm]{p1.pdf}}
%\caption{Flowchart of the proposed Weak-supervised Cross-modal Hashing. It encompasses three steps: 1. Enriching the label matrix via weakly multi-label leaning using weakly multi-label information and multi-modal training data; 2. Converting the multi-modal data from different modalities onto the same dimensional representation and optimizing a central modality; 3. Learning hashing functions for all modalities by leveraging the enriched label matrix with respect to the central modality.}
%\label{Fig1}
%\end{figure*}

The remainder of this paper is organized as follows. We briefly review  representative multi-modal hashing and zero-shot hashing methods in Section \ref{Related Work},  and then elaborate on the proposed algorithm and its optimization in Section \ref{Cross}. Section \ref{sec:exp} provides  experimental results and analysis on three real-world datasets. Conclusions and future work are given in Section \ref{sec:concl}.

\section{Related Work}
Our work is closely related to cross-modal hashing and zero-shot hashing (as illustrated in Table \ref{Table1}). Cross-modal zero-shot hashing can be viewed as a special case of zero-shot learning and hashing learning. Interested readers can refer to \cite{wang2016L2H,wang2018L2H} and \cite{xian2018zslsurvey,wang2019zslsurvey}. In the following, we give a brief review of related work, and emphasize how our approach differs.

\label{Related Work}

\begin{table}[h!tbp]
\label{Table1}
\scriptsize
\centering
\caption{Multiplicity data handled by different hashing methods. 'un', 'su' and 'se' are short for 'unsupervised', 'supervised' and 'semi-supervised' CMH, respectively.}
\begin{tabular}{c|c|c|c|c|c|c}
\hline
\multicolumn{2}{c}{Category}  &  Methods & \tabincell{c}{Multi-\\modal} & \tabincell{c}{Unseen\\class} & \tabincell{c}{Unlabeled\\ data} & \tabincell{c}{Different\\ label spaces}\\\hline

\multirow{9}{*}{\tabincell{c}{CMH}}
&\tabincell{c}{un} & \tabincell{c}{CVH\cite{Kumar2011Learning}\\CMFH\cite{Ding2014Collective}} & \tabincell{c}{$\surd$\\$\surd$} & \tabincell{c}{$\times$\\$\times$} &
\tabincell{c}{$\times$\\$\times$} &
\tabincell{c}{$\times$\\$\times$} \\\cline{2-7}

& \tabincell{c}{su} & \tabincell{c}{CMSSH\cite{Bronstein2010Data}\\SCM\cite{Lin2015Semantics}\\QCH\cite{Wu2015Quantized}\\SePH\cite{Lin2015Semantics}\\DCMH\cite{Jiang2017Deep}\\PRDH\cite{yang2017pairwise}\\AgNet \cite{Zhong2018Attribute}} & \tabincell{c}{$\surd$\\$\surd$\\$\surd$\\$\surd$\\$\surd$\\$\surd$\\$\surd$} & \tabincell{c}{$\times$\\$\times$\\$\times$\\$\times$\\$\times$\\$\times$\\$\surd$} &
\tabincell{c}{$\times$\\$\times$\\$\times$\\$\times$\\$\times$\\$\times$\\$\times$} &
\tabincell{c}{$\times$\\$\times$\\$\times$\\$\times$\\$\times$\\$\times$\\$\times$} \\\cline{2-7}

& \tabincell{c}{se} & \tabincell{c}{RDCMH\cite{liu2019ranking}\\\textbf{CZHash}} & \tabincell{c}{$\surd$\\$\surd$} & \tabincell{c}{$\times$\\$\surd$} &
\tabincell{c}{$\surd$\\$\surd$} &
\tabincell{c}{$\times$\\$\surd$} \\\hline

\tabincell{c}{ZSH} & & \tabincell{c}{TSK\cite{Yang2016Zero}\\BZSL\cite{shen2019scalable}} & \tabincell{c}{$\times$\\$\times$} & \tabincell{c}{$\surd$\\$\surd$} & \tabincell{c}{$\times$\\$\times$} & \tabincell{c}{$\times$\\$\times$}\\\hline
%\multicolumn{2}{c}{\tabincell{c}{Cros-modal\\ zero-shot hashing}} & \tabincell{c}{AgNet \cite{Zhong2018Attribute}\\CZHash} & \tabincell{c}{$\surd$\\ $\surd$} & \tabincell{c}{$\surd$\\ $\surd$} & \tabincell{c}{$\times$\\ $\surd$}& \tabincell{c}{$\times$\\ $\surd$}\\
\hline
\end{tabular}\\
\label{Table1}
\end{table}

%\subsection{Cross-modal Hashing}
%Cross-modal hashing has attracted increasing interest in various domains in recent years. A comprehensive coverage of these methods is out of scope for this paper, and we only  review some of the most related and representative cross-modal hashing methods. Excellent surveys exist for interested readers \cite{wang2016L2H,wang2018L2H}. Existing cross-modal hashing methods can be divided into two main categories: handcraft-based and deep-based.
%The former used the handcraft features and the latter use features extracted by the deep neural network.

%{\color{blue}Organized from unsupervised, supervised, semi, and then transfer to deep learning (1st paragraph). After that, move to zero shot hashing (second paragraph). Finally drive out our solution (3rd paragraph)}

%typically use the features extracted by classical hand-crafted methods such as Scale-invariant feature transform(SIFT) and Latent Dirichlet Allocation(LDA).

%Latent semantic sparse hashing (LSSH)\cite{Zhou2014Latent} utilizes sparse coding for images and matrix factorization for text  to learn the latent semantic features and eventually map the learned features onto a joint abstraction space to generate unified hash codes.
\textbf{Cross-modal Hashing}: Existing cross-modal hashing can be divided into three categories: unsupervised, supervised, and semi-supervised. Unsupervised methods typically  seek
hash coding functions by taking into account underlying data distributions or correlations. To name a few, Cross-view hashing (CVH) extends the single-view spectral hashing to multi-modal scenario by mapping
similar objects to similar codes across views to enable cross-view similarity search.
%\{{\color{red}{{Please fix the previous sentence. It does not make sense. What does it mean similarity-weighted distances??? CD}}}\}
\cite{Kumar2011Learning}. Collective matrix factorization hashing (CMFH) learns unified hash codes by establishing a latent factor model across different modalities via collective matrix factorization \cite{
Ding2014Collective}. On the other hand, supervised cross-modal hashing use semantic labels (or ranking order) of training data, to improve the performance. For example, cross-modal similarity sensitive hashing (CMSSH) \cite{Bronstein2010Data} models the projections to hash codes as binary classification problems with positive and negative examples using features in each view, and utilizes a boosting method to efficiently learn the hash functions. Semantic correlation maximization (SCM) \cite{Zhang2014Large} optimizes the hashing functions by maximizing the correlation between two modalities with respect to the semantic labels. Quantized Correlation Hashing (QCH) \cite{Wu2015Quantized}  simultaneously optimizes the quality loss over all data modalities and the pairwise relationships between modalities.
Semantics Preserving Hashing (SePH) \cite{Lin2017Cross} generates one unified hash code for all observed views of any instance using the semantic consistency between views.

Recently, deep learning has also been incorporated with cross-modal hashing. Deep cross-modal hashing (DCMH) \cite{Jiang2017Deep} combines hashing learning and deep feature learning by preserving the semantic similarity between modalities. Correlation auto-encoder hashing (CAH) \cite{Cao2016Correlation}  embeds the maximum cross-modal similarity into hash codes using nonlinear deep autoencoders. Correlation hashing network (CHN) \cite{Yue2016Correlation}  jointly learns image and text representations tailored to hash coding and formally controls the quantization error.
Pairwise relationship guided deep hashing (PRDH) \cite{yang2017pairwise} jointly uses two types of pairwise constraints from intra-modality and inter-modality to preserve the semantic similarity of the learned hash codes.
Semi-supervised cross-modal hashing leverages labeled and unlabeled data to train the hashing model. For example, ranking-based deep cross-modal hashing (RDCMH) \cite{liu2019ranking} can  preserve multi-level semantic similarity between labeled and unlabeled multi-label objects for cross-modal hashing.

\textbf{Zero-shot Hashing}: ZSH can be viewed as a combination of ZSL and hashing-based retrieval techniques. It can handle the close-set limitation in hashing-based retrieval approaches, i.e., the concepts of possible testing instances, either in the database or in query modality, are \emph{not in} the training set \cite{xian2018zslsurvey}. Therefore, ZSH must exploit information from seen categories to build hash functions and to retrieve the samples in unseen categories.  For example, Transferring Supervised Knowledge (TSK) \cite{Yang2016Zero} takes the semantic vectors obtained by embedding labels as a bridge to transfer available supervision information from seen
categories to unseen categories. Binary embedding based Zero-Shot Learning (BZSL) \cite{shen2019scalable} jointly learns two binary coding functions  to map class embeddings and visual features into the binary embedding space, which could well preserve the discriminative category. AgNet \cite{Zhong2018Attribute} is the first (to the best of our knowledge) cross-modal zero-shot hashing method, which aligns different modality data in a semantically enriched attribute space, and  uses the attribute vectors to guide the generation of hash codes for image and text modalities within the same network. However, AgNet trains the deep neural network to learn fine-grain image and text features separately, and then uses the combined features as the input of hashing quantification networks. This may lead to incompatible  feature learning  and  quantification processes. Furthermore, AgNet is designed for cross-modal zero-shot single-labeled data, where each instance is tagged with only one class label. As a result, AgNet cannot be used with multi-label datasets, in which each instance can be tagged with a set of non-exclusive labels.

%Besides, all these ZSH methods utilize the label information by embedding the labels into the semantic embedding space,  which may perform weakly when dealing with few labeled but abundant unlabeled data. Moreover, these  methods can not handle multi-modality data with associated with different label spaces.

Unlike existing CMH and ZSH solutions, our CZHash has to handle three open but general \textbf{problems}: (i) new category labels (zero-shot), (ii) labeled and unlabeled data, and  (iii) training data with different label spaces. The first problem implies that the supervision knowledge is limited to seen categories, which is the only information available during the learning of reliable hash functions for transforming samples of unseen categories into binary codes. The second problem implies that the labeled instances of training data is not sufficient, where few are labeled and many are unlabeled. The third problem implies that the label spaces of different data modalities may also be different. The distinctions between our CZHash and existing cross-modal hashing and zero-shot hashing solutions are listed in Table \ref{Table1}. Clearly, our CZHash approach is more challenging and general than the existing solutions.

\section{The Proposed Method}
\label{Cross}
\subsection{Notation and Problem Definition}
The notation used in this paper is given in Table \ref{Table2}. Without loss of generality, we assume that the first $l$ samples are annotated with category labels (which may be different across modalities), and the remaining $u=n-l$ samples are unlabeled. To formulate the problem, we kick off the cross-modal hashing with $m=2$, but our solution can easily adapt to  $m\geq 3$ modalities. For the zero-shot setting, there are $p_1$ ($p_2$) unseen category labels (or new concepts) for the two modalities respectively. Each category label $h$ in the different modalities is also represented by an attribute vector, $\mathbf{a}^{(1)}_h$  and $\mathbf{a}^{(2)}_h$, which  captures the latent relationship between seen and unseen categories. CZHash aims to learn two hashing functions, $\{H_1\}$: $\mathbb{R}^{d_{1}}\rightarrow {\{0,1}\}^{b}$ and $\{H_2\}$: $\mathbb{R}^{d_{2}}\rightarrow {\{0,1}\}^{b}$,  which can map the feature vectors in the respective modality onto a common Hamming space, and use the attribute vectors to explore the correlations from seen labels to unseen ones, while preserving the proximity of the original data. We observe that, unlike existing ZSH solutions and AgNet, CZHash can leverage both labeled and unlabeled data, and account for multiple modalities annotated with different sets of labels.

\begin{comment}
Suppose $\mathbf{X}^{(v)}={\{\mathbf{x}^{(v)}_1, \mathbf{x}^{(v)}_2, \cdots, \mathbf{x}^{(v)}_n}\} \in \mathbb{R}^{n \times d_m}$  are $m$ data modalities, $n$ is the number of instances (data points), $d_m$ is the dimensionality of the instances in the respective modality. %For example, in the Wiki-image search application, $x^{(1)}_i$ is the image features of the entity $i$, and $x^{(2)}_i$ is the text features of this entity.
$\mathbf{L}^{(v)} \in \mathbb{R}^{n \times q^{(v)}}$  store the different label information of $n$ instances in $\mathbf{X}^{(v)}$  respect to $q_m$ distinct labels in the seen category label sets $\mathcal{L}^{(v)}$. %$\mathbf{L}^{(v)}_(i,k)\in{\{0,1}\}$, $\mathbf{L}^{(v)}_(i,k)=1$ indicates that $x^{(v)}_i$ is labeled with the $k$-th label; $\mathbf{L}^{(v)}_(i,k)=0$ otherwise.

\end{comment}
%{\color{red}[PLEASE short the above paragraphs to inline with the table.]}
%{\color{red}[This time we can summarize a symbol table!]}

\begin{table}
\label{Table2}
\centering
\caption{Used Notation.}
\begin{tabular}{l|l}
\hline

 $b$ & hashing code length\\
 $c$ & number of categories\\
 $d$ & size of attribute vectors\\
 $n$ & number of instances\\
 $l$ & number of labeled instances\\
 $u$ & number of unlabeled instances\\
% $m$ & number of modalities\\
 %$p_1$ and $p_2$& number of unseen classes of the two modalities\\
 $q_1$ and $q_2$ & number of seen labels of two modalities\\
 $\mathbf{x}^{(1)}_i \in \mathbb{R}^{d_1}$, $\mathbf{x}^{(2)}_i \in \mathbb{R}^{d_2}$  & feature vectors in the two modalities\\
 $\mathcal{L}^{(1)}_i \in \mathbb{R}^{q_1}$, $\mathcal{L}^{(2)}_i \in \mathbb{R}^{q_2}$  & seen label set of $\mathbf{x}^{(1)}_i$ and  $\mathbf{x}^{(2)}_i$\\
 $\mathbf{a}^{(1)}_h \in \mathbb{R}^{d}$, $\mathbf{a}^{(2)}_h \in \mathbb{R}^{d}$  &attribute vectors of the $h$-th category\\
 $\mathbf{F}^{(1)},  \mathbf{F}^{(2)} \in \mathbb{R}^{n \times d} $  &deep feature matrix for the $v$-th modality \\
 $\mathbf{S}^{11}, \mathbf{S}^{22}, \mathbf{S}^{12} \in \mathbb{R}^{n\times n}$, &intra-modal  or inter-modal  similarity matrices\\
 $\mathbf{C}^{(1)},  \mathbf{C}^{(2)} \in \mathbb{R}^{n \times c}$ &  category attribute matrices\\
 $\mathbf{W}^{(1)}, \mathbf{W}^{(2)} \in \mathbb{R}^{c \times b}$  & coefficient matrix\\
 $\mathbf{B} \in \mathbb{R}^{n \times b}$  & hashing code matrix\\
\hline
\end{tabular}\\
\label{Table2}
\end{table}

CZHash involves two main steps. It firstly quantifies the composite similarity between instances within (or between) modalities, based on the label and feature information with respect to both labeled and unlabeled samples. Then, it defines an objective function to simultaneously achieve deep feature learning, category space learning, and hashing coding function learning under the guidance of the composite similarity.  The overall workflow of CZHash is shown in Fig. \ref{Fig1}.
\vspace{-2pt}
\begin{figure*}[h!t]
\centering
{\label{Fig1}\includegraphics[width=18cm,height=6.2cm]{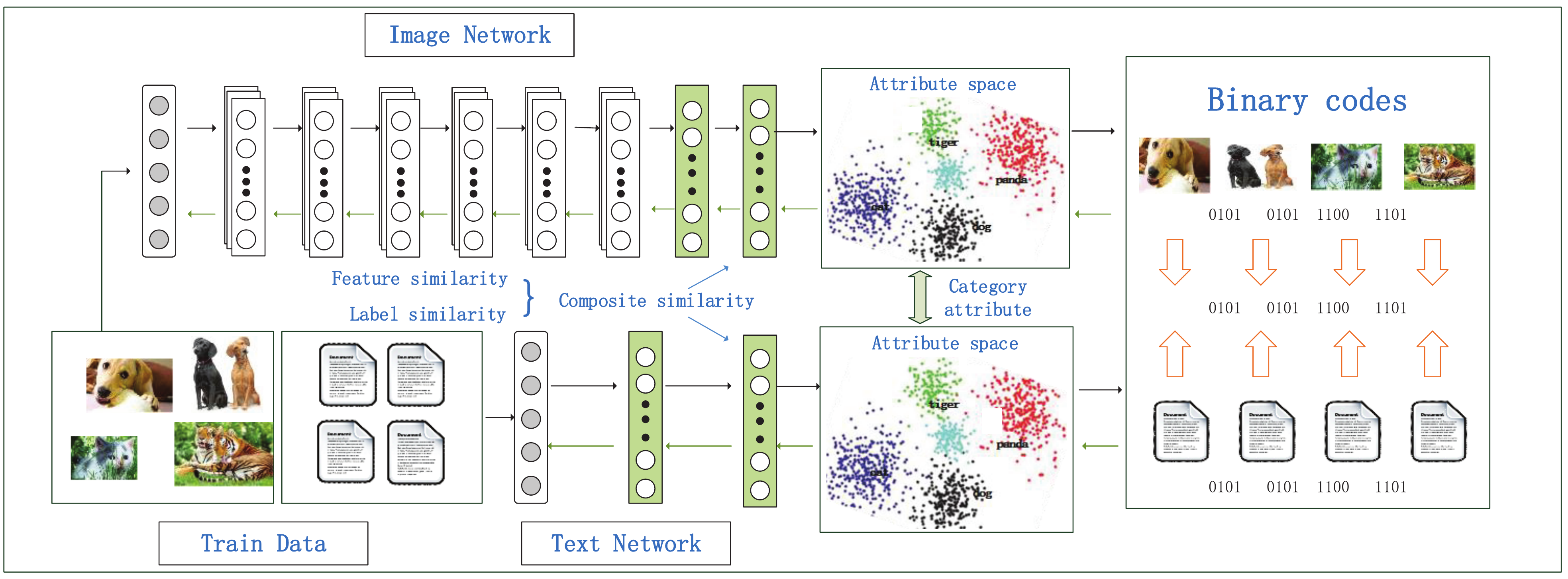}}
\caption{The architecture of the proposed   Cross-Modal Zero-shot Hashing (CZHash). CZHash includes four steps: (1) learning feature representations with the proposed composite similarity via an image CNN network and a text two-layer network; (2) constructing the category attribute spaces for each modality using the learned deep features; (3) obtaining the binary hashing codes based on the category attribute spaces and quantification hashing functions; (4) simultaneously optimizing deep features, compatible with semantic similarity preserving, category attribute space learning, and hashing quantification learning, via a unified objective function.}
\label{Fig1}
\end{figure*}
\vspace{-2pt}
\subsection{Semi-supervised Composite Similarity}
\vspace{-1pt}
Supervised cross-modal hashing methods achieve favorable performance, since they utilize intrinsic properties (i.e., the labels) of samples, along with the structural information, to guide the hashing codes learning and the cross-modal similarity preserving. Most existing supervised hashing methods \cite{Bronstein2010Data,Zhang2014Large,Lin2017Cross} typically quantifies the pairwise semantic similarity between instances using the cosine similarity between the semantic label vectors of respective samples. However, since the label spaces ($\mathcal{L}^{(1)} \in \{0|1\}_{1}^{q_1}$ and $\mathcal{L}^{(2)} \in \{0|1\}_{1}^{q_2}$) of different modalities are not the same ($q_1 \neq q_2$) in our general cross-modal zero-shot hashing setting, cosine similarity cannot directly apply to our situation. Given this, we measure the semantic similarity with the Jaccard similarity as follows:
\begin{equation}
\hat{s}_{ij}^{vv'}=\frac{|\mathcal{L}_i^{v} \cap \mathcal{L}_{j}^{v'}|}{|\mathcal{L}_i^{v} \cup \mathcal{L}_j^{v'}|}, \{v, v' \in \{1, 2\}\}
\label{Eq1}
\end{equation}
where $|\mathcal{L}_i^{v}|$ is the cardinality of $\mathcal{L}_i^{v}$, which corresponds to the number of labels assigned to the $i$-th instance in the $v$-th ($v\in \{1,2\}$) modality. We want to remark that other set based similarity metrics can also be adopted to quantify $\hat{s}_{ij}^{vv'}$, which is not the main focus of our work. However, Eq. (\ref{Eq1}), like cosine similarity, can only capture the semantic similarity between samples with labels. Sufficient labeled samples are expected to enable credible quantification and zero-shot hashing. On the other hand, only limited labeled instances, and abundant unlabeled instances, are available.

To remedy the issue of insufficient labeled data, we introduce a composite similarity measure that takes into account both the label and feature information of training data. The labels of an instance depend on its features, and the semantic similarity is positively correlated with the feature similarity of the respective instances \cite{zhang2010multilabel,wang2009multi}. As such, we can integrate the feature similarity between data with their semantic similarity.  As a result, we define a \emph{semi-supervised composite similarity} as follows:
\vspace{-1pt}
\begin{equation}
{s}_{ij}^{11}=\left\{
             \begin{array}{lr}
            \check{s}_{ij}^{11}{(1+(\hat{s}_{ij}^{11}-\check{s}_{ij}^{11}))}, & \mathcal{L}^{1}_i\neq {\emptyset}\ \ \textrm{and} \ \ \mathcal{L}^{1}_j\neq {\emptyset}  \\
             \check{s}_{ij}^{11}, & \textrm{otherwise}
             \end{array}
\right.
\label{Eq2}
\end{equation}
where $\check{s}_{ij}^{11}$ is the within modality feature similarity and $\hat{s}_{ij}^{11}$ is the semantic similarity between $\mathbf{x}^{(1)}_i$ and $\mathbf{x}^{(1)}_j$, respectively. The former is computed by the Euclidean similarity ($1/(1+edist(i,j))$, where $edist(i,j)$ is the Euclidean distance between  $\mathbf{x}^{(1)}_i$ and $\mathbf{x}^{(1)}_j$), other similarity metrics can also be used. The latter is calculated by Eq. (\ref{Eq1}). Note, ${s}_{ij}^{11}, \check{s}_{ij}^{11}$ and $\hat{s}_{ij}^{11}$ are always in the interval [0,1]. Eq. (\ref{Eq2}) can account for both  labeled and unlabeled training data. Specifically, for two unlabeled data, the composite similarity between  $\mathbf{x}^{(1)}_i$ and $\mathbf{x}^{(1)}_j$ is directly computed from feature information. For labeled data, we assume that the label similarity  $\hat{s}_{ij}^{11}$ is a supplement for $\check{s}_{ij}^{11}$. The larger $\hat{s}_{ij}^{11}$ is, the larger ${s}_{ij}^{11}$ will be. In this way, we can not only leverage the label and feature information of training data to account for insufficient labels, but also retrieve semantic related instances. Similarly to $\mathbf{S}^{11}\in \mathbb{R}^{n \times n}$, we can obtain the intra-modality composite similarity matrix  $\mathbf{S}^{22}\in \mathbb{R}^{n \times n}$ from the other data modality.

To perform cross-modal hashing, we should use the semantic similarity and pairing relationships between instances across modalities. For the inter-modality relation matrix $\mathbf{S}^{12}\in \mathbb{R}^{n \times n}$, since instances of different modalities have heterogeneous features, we cannot measure  feature similarity directly in this case as done within a modality. As such, we quantify the composite similarity between two instances of different modalities as follows:
\begin{equation}
{s}_{ij}^{12}=\left\{
             \begin{array}{lr}
           \hat{s}^{12}_{ij}(1+( ({s}^{11}_{ij}+{s}^{22}_{ij})/2)-\hat{s}^{12}_{ij}), & \mathcal{L}^1_i\neq {\emptyset}\ , \mathcal{L}^2_j\neq {\emptyset}  \\
             1/2({s}^{11}_{ij}+{s}^{22}_{ij}), & \textrm{otherwise}
             \end{array}
\right.
\label{Eq3}
\end{equation}
where $\hat{s}^{12}_{ij}$ is calculated by Eq. (\ref{Eq1}). Here, the feature similarity between two instances of different modalities is the average of their paired instances within the same modality. This average can balance the structural inconsistency across modalities since the  similarity of same instances between different modalites may exist semantic gap due to the different construction and label spaces. In addition, when two instances across modalities have overlapping labels, they should also be semantically related. As such, similarly to Eq. (\ref{Eq2}), the semantic similarity between labeled instances is also considered.

\subsection{Cross-Modal Hashing}
\subsubsection{Deep Feature Representation}
Given the promising performance of deep neural networks, and their ability to represent complex data, here we design deep hash functions using CNN \cite{krizhevsky2012cnn} to jointly learn feature representations, and their mappings to category spaces and hash codes. A non-liner hierarchical hash function is more suitable for feature representation to preserve the cross-modal similarity construction. We emphasize that  deep neural network design is not the main focus of our work, and other representation learning models (i.e., AlexNet) can  be used to learn deep features of images and text for CZHash. The deep feature learning process mainly involves two parts: an image-network and a text-network.

The adopted deep neural network for the image modality includes eight layers: the first six layers are the same as those in CNN-F \cite{Chatfield2014Return}; the seventh and eighth layers are fully-connected, with the output being the learned image features.
We first represent the text as a bag-of-words (BOW) vector. We then input the BOW vectors to a neural network with two fully-connected layers, and take the outputs as the learned text features. The details of each layer in the image and text networks are shown in Table \ref{Table7}; `filters' and `stride' represent the sizes of the convolution filters and of the convolution stride, respectively; `nodes'  denotes the number of nodes in the fully connected layers, and it is also the dimensionality of the output at that layer; `relu' and `tanh' are the adopted activation functions in each layer.

\begin{table}
\label{Table7}
\centering
\caption{Data types handled by different hashing methods.}
\begin{tabular}{c|c|c}
\hline
Network  &  Layers & Configuration \\\hline
Image-network & \tabincell{c}{conv1\\conv2\\conv3\\conv4\\conv5\\conv6\\full1\\full2} & \tabincell{c}{filters:$64\times11\times11$, stride:$4\times4$\\filters:$265\times5\times5$, stride:$1\times1$\\filters:$265\times5\times5$, stride:$1\times1$\\filters:$265\times3\times3$, stride:$1\times1$\\filters:$265\times3\times3$, stride:$1\times1$\\filters:$265\times3\times3$, stride:$1\times1$\\nodes:$4096$, relu\\nodes:$d$, tanh}  \\\hline
Text-network & \tabincell{c}{full1\\full2} &  \tabincell{c}{nodes:4096, relu\\nodes:$d$, tanh}\\
\hline
\end{tabular}\\
\label{Table7}
\end{table}

In the following, we denote the learned deep feature representations of $x^{(1)}$ and $x^{(2)}$ as $\varphi(x^{(1)})$ and $\phi(x^{(2)})$. The parameter $\theta_x^{(1)}$ and $\theta_x^{(1)} $ optimization of the non-linear mappings for the two representations is discussed in the next section. Specifically, we utilize the composite similarity obtained in the previous subsection to guide the deep feature learning as follows:
\begin{equation}
\begin{split}
\mathop {\min }\limits_{\mathbf{F}^{(1)},\mathbf{F}^{(2)}} &||{\mathbf{F}^{(1)}\mathbf{F}^{(1)}}^T-\mathbf{S}^{11}||^2_F+||{\mathbf{F}^{(2)}\mathbf{F}^{(2)}}^T-\mathbf{S}^{22}||^2_F\\&+2||{\mathbf{F}^{(2)}\mathbf{F}^{(1)}}^T-\mathbf{S}^{12}||^2_F
\end{split}
\label{Eq4}
\end{equation}
where $\mathbf{F}^{(1)}_{*i}=(\varphi(x^{(1)}_i);\theta_x^{(1)})$ and $\mathbf{F}^{(2)}_{*i}=(\phi(x^{(2)}_i);\theta_x^{(2)})$. The first and second terms of Eq. (\ref{Eq4}) aim to preserve the intra-modality composite similarity of image and text, respectively,  and the third term pursues the inter-modality similarity preservation.

\subsubsection{Category Space learning}
After obtaining the deep feature representations of different modalities, we aim to learn an embedding category space with the attribute vectors as guidance to achieve zero-shot hashing. To learn the unseen classes from the seen ones, previous zero-shot hashing methods (i.e., TSK and BZSL) typically embed the whole label space,  or the instance-level attributes (as AgNet does using complete labels), into the category space to explore the correlations between seen and unseen classes. As a result, they may perform poorly when sufficient labeled instances are not available. Given this, ZCHash leverages class-level attribute vectors, which just require the  name of classes without demanding complete labels, to guide the deep feature mapping into the category space.
Specifically, we combine the learned deep features with the category attribute vector to obtain the semantic category representation  with respect to each category for instances. With the learned category representation vector for each instance, the relationships between seen and unseen classes can be well captured and characterized. To  achieve cross-modal zero-shot hashing  and to avoid domain differences, semantic shift and  semantic  variation, we optimize the category representation space as follows:
\begin{equation}
\begin{split}
\mathop {\min }\limits_{\mathbf{C}^{(1)},\mathbf{C}^{(2)}} &||\mathbf{F}^{(1)}-\mathbf{C}^{(1)}\mathbf{A}^{(1)}||^2_F+||\mathbf{F}^{(2)}-\mathbf{C}^{(2)}\mathbf{A}^{(2)}||^2_F
%+\\&||R^T\mathbf{C}^{(1)}-\mathbf{C}^{(2)}||^2_F
\end{split}
\label{Eq5}
\end{equation}
where $\mathbf{C}^{(1)}, \mathbf{C}^{(2)} \in \mathbb{R}^{n\times c}$  are the learned semantic category spaces for each instance of the image and text modalities, respectively. $\mathbf{A}^{(1)}$ and $\mathbf{A}^{(2)} \in \mathbb{R}^{c\times d}$ are the attribute matrices, which define the correlations between categories.  Since  different modalities are in essence from the same domain, the latent categories from different modalities can be the same, although the observed category labels may be different. As a result, without loss of generality, both $\mathbf{A}^{(1)}$ and $\mathbf{A}^{(2)}$ are set to their union.

\subsubsection{Cross-modal Hashing}
Finally,  we can seek the hashing codes from the learned category attribute spaces by minimizing the hashing quantification loss as follows:
\begin{equation}
\begin{split}
\mathop {\min }\limits_{\mathbf{B}} ||\mathbf{C}^{(1)}\mathbf{W}^{(1)}-\mathbf{B}^{(1)}||^2_F+||\mathbf{C}^{(2)}\mathbf{W}^{(2)}-\mathbf{B}^{(2)}||^2_F
\end{split}
\label{Eq6}
\end{equation}
where $\mathbf{W}^{(1)}$ and $\mathbf{W}^{(2)} \in \mathbb{R}^{c\times b}$ are the coefficient matrices of the two modalities. $\mathbf{B}^{(1)}$ and $\mathbf{B}^{(2)} \in \mathbb{R}^{n\times b}$ represent the binary code matrices of the image and text modalities. In the training process, since different modality data of the same sample actually describe the sample from different viewpoints, we can set the binary codes of the same training instances from two modalities as the same, namely $\mathbf{B}^{(1)}=\mathbf{B}^{(2)}=\mathbf{B}$.

To this end,  we can integrate the deep feature learning process with  composite similarity preserving,  category space learning, and the hashing quantification process, and form a unified objective function as follows:
\begin{equation}
\begin{split}
&\mathop {\min }\limits_{{\substack{\mathbf{F}^{(v)},\mathbf{C}^{(v)},\\ \mathbf{W}^{(v)},\mathbf{B}}}} \mathbf{J}_{loss}= \sum_{v,v'=1}^2||{{\mathbf{F}^{(v)}}\mathbf{F}^{(v')}}^T-\mathbf{S}^{vv'}||^2_F\\& +\alpha\sum_{v=1}^2||\mathbf{F}^{(v)}-\mathbf{C}^{(v)}\mathbf{A}^{(v)}||^2_F
%\\&+\sum_{\substack{k=1\\m=k+1}}^M||{\mathbf{R}^{(v)}}^T\mathbf{C}^{(k)}-\mathbf{C}^{(v)}||^2_F)
+\beta\sum_{v=1}^2||\mathbf{C}^{(v)}\mathbf{W}^{(v)}-\mathbf{B}||^2_F
\end{split}
\label{Eq7}
\end{equation}
where the first term measures the loss due to the deep representations with respect to the corresponding similarity matrices; the second term measures the loss incurred by the semantic category spaces; and the third term defines the hashing quantification loss. $\alpha$ and $\beta$ are the hyper-parameters.

\subsubsection{Optimization}
We can solve Eq. (\ref{Eq7}) via the Alternating Direction Method of Multipliers (ADMM) \cite{Boyd2011Distributed},
which alternatively optimizes one of $\textbf{F}^{(v)}$, $\textbf{C}^{(v)}$,   $\mathbf{W}^{(v)}$, and $\textbf{B}$, while keeping the other three fixed. Eq. (\ref{Eq7}) can be optimized as follows.

\textbf{Optimize $\textbf{F}^{(v)}$}: By fixing all variables but $\textbf{F}^{(1)}$,  for each sampled instance, we first compute the following gradient:
\begin{equation}
%\scriptsize
\begin{split}
\frac{\partial \mathbf{J}_{loss}}{\partial \mathbf{F}^{(1)}_{*i}} =& \frac{1}{2} (\mathbf{S}^{11}_{*i}+\mathbf{S}^{22}_{*i})\mathbf{F}^{(1)}_{*i}-\mathbf{S}^{12}_{*i}\mathbf{F}^{(2)}_{*i}+\alpha(2\mathbf{F}^{(1)}_{*i}-\mathbf{C}^{(1)}_{*i}\mathbf{A}^{(1)})
\end{split}
\label{Eq8}
\end{equation}
The derivative with respect to $\mathbf{F}^{(2)}_{*i}$ is similar.

\textbf{Optimize $\textbf{C}^{(v)}$}: By fixing all variables but $\textbf{C}^{(1)}$, we can obtain:
\begin{equation}
\begin{split}
\frac{\partial \mathbf{J}_{loss}}{\partial \mathbf{C}^{(1)}_{*i}} = &\frac{1}{2}\alpha\mathbf{F}^{(1)}_{*i}{\mathbf{A}^{(1)}}^T
+\beta(\mathbf{C}^{(1)}_{*i}\mathbf{W}^{(1)}-\mathbf{B}_{*i}){\mathbf{W}^{(1)}}^T
\end{split}
\label{Eq9}
\end{equation}
The derivative with respect to $\mathbf{C}^{(2)}_{*i}$ is similar.
%\begin{equation}
%\begin{split}
%\frac{\partial L}{\partial \mathbf{C}^{(2)}_{*i}} = &\frac{1}{2}\alpha(2\mathbf{F}^{(2)}_{*i}{\mathbf{A}^{(2)}}^T_{*i}-2R^T_{*i}\mathbf{C}^{(1)}_{*i}+2\mathbf{C}^{(2)}_{*i})\\&+\beta(\mathbf{C}^{(2)}\mathbf{W}^{(2)}_{*i}-B_{*i}){\mathbf{W}^{(2)}}^T_{*i}
%\end{split}
%\label{eq8}
%\end{equation}

%\textbf{Optimize $\textbf{R}$}: By fixing all other variables but $\textbf{R}$, we can get:
%\begin{equation}
%\begin{split}
%\frac{\partial L}{\partial \mathbf{R}_{*i}}=R_{*i}\mathbf{C}^{(1)}_{*i}C^{{(1)}^T}_{*i}-2\mathbf{C}^{(2)}_{*i}C^{{(1)}^T}_{*i}
%\end{split}
%\label{eq9}
%\end{equation}

\textbf{Optimize $\mathbf{W}^{(v)}$}: By fixing all variables but $\mathbf{W}^{(1)}$, we can get:
\begin{equation}
\begin{split}
\frac{\partial \mathbf{J}_{loss}}{\partial \mathbf{W}^{(1)}}={\mathbf{C}^{(1)}}^T\mathbf{C}^{(1)}\mathbf{W}^{(1)}-{\mathbf{C}^{(1)}}^T\mathbf{B}^T
\end{split}
\label{Eq10}
\end{equation}
The derivative with respect to $\mathbf{W}^{(2)}$ is again similar.

Like most of the existing deep learning methods, we
use stochastic gradient descent (SGD) and
the back-propagation (BP) algorithm to learn the parameters of the deep neural network. Specifically, we can compute the parameters  in the first modality with $\frac{\partial \mathbf{J}_{loss}}{\partial \mathbf{F}^{(1)}_{*i}}$, $\frac{\partial \mathbf{J}_{loss}}{\partial \mathbf{C}^{(1)}_{*i}}$ and $\frac{\partial \mathbf{J}_{loss}}{\partial \mathbf{W}^{(1)}}$  using the chain rule, and proceed similarly in the other modalities. These derivative values are used to update the hashing code matrix $\mathbf{B}$, which is then fed into different layers of the CNN to update the parameters of $\varphi(x^{(1)})$ ($\varphi(x^{(2)})$) in each layer via the BP algorithm.

\textbf{Optimize $\textbf{B}$}: Once $\textbf{F}^{(v)}$, $\textbf{C}^{(v)}$,  and $\textbf{W}^{(v)}$  are optimized and fixed, the parameters of the deep neural network are learned. The minimization problem in Eq. (\ref{Eq7}) is transformed into a maximization problem (given the fixed $\textbf{C}^{(v)}$ and $\textbf{W}^{(v)}$) as follows:
\begin{flalign}
%\scriptsize
\label{Eq11}
\begin{split}
\mathop {\max }\limits_{\mathbf{B}} tr(\lambda \mathbf{B}^T(\mathbf{C}^{(1)}\mathbf{W}^{(1)}\!+\!\mathbf{C}^{(2)}\mathbf{W}^{(2)}))\!=\!tr(\mathbf{B}^T\mathbf{U})\!=\!\sum_{i,j}\mathbf{B}_{ij}\mathbf{U}_{ij}
\end{split}
\end{flalign}
where $\mathbf{B}\in{\{1,-1}\}^{n\times c}, \mathbf{U}={\lambda}(\mathbf{C}^{(1)}\mathbf{W}^{(1)}+\mathbf{C}^{(2)}\mathbf{W}^{(2)})$. It's easy to see that the binary code $\mathbf{B}_{ij}$ should have the
same sign as $\mathbf{U}_{ij}$ for the maximization. Therefore, we have:
\begin{equation}
\label{Eq12}
\mathbf{B}=sign(\mathbf{U})=sign({\lambda}(\mathbf{C}^{(1)}\mathbf{W}^{(1)}+\mathbf{C}^{(2)}\mathbf{W}^{(2)}))
\end{equation}

The optimal $\textbf{F}^{(v)}$, $\textbf{C}^{(v)}$,  $\textbf{W}^{(v)}$, and $\textbf{B}$ can be iteratively optimized via  Eqs. (\ref{Eq7}-\ref{Eq12}). Using the minibatch stochastic gradient descent algorithm, we set the mini-batch size for gradient descent to 128, and the dropout rate to 0.5 on the fully connected layers to avoid overfitting.

\section{Experiments}
In this Section, we conduct a set of experiments to study the effectiveness of CZHash under different scenarios of cross-modal hashing, and compare its performance against  related  methods.
\label{sec:exp}
\subsection{Experimental setup}
\textbf{Datasets}:
We conduct experiments on three  benchmark datasets, Nus-wide, Wiki, and Mirflicker. The modalities of these datasets consist of images and texts, but CZHash can also be applied  to $m \geq 3$ modalities.

Nus-wide{\footnote{http://lms.comp.nus.edu.sg/research/NUS-WIDE.htm}} is a large-scale dataset crawled from Flickr. It contains 260,648 labeled images associated with user tags, which can be considered as image-text pairs. Each image is annotated with one or more labels taken from 81 concept labels. Each text is represented as a 1,000-dimensional bag-of-words vector. The hand-crafted features of each image correspond to a 500-dimensional bag-of-visual word (BOVW) vector.
Wiki{\footnote{https://www.wikidata.org/wiki/Wikidata}} is generated from a group of 2,866 Wikipedia documents. Each document is an image-text pair. In each pair, the text is an article describing people, places or some events, and the image is closely related to the content of the article. Each image can be annotated with 10 semantic labels, and is represented by a 128-dimensional SIFT feature vector for hand-crafted features. The text articles are represented as probability distributions over 10 topics, which are derived from a Latent Dirichlet Allocation (LDA) model.
Mirflickr{\footnote{http://press.liacs.nl/mirflickr/mirdownload.html}} originally includes 25,000 instances collected from Flicker. Each instance consists of an image and its associated textual tags, and is manually annotated with one or more labels, from a total of 24  semantic labels. For  hand-crafted features, each text is represented as a 1,386-dimensional bag-of-word vector, and each image is represented by a 512-dimensional GIST feature vector.
%This dataset can be downloaded from %\href{http://press.liacs.nl/mirflickr/mirdownload.html}

For the zero-shot settings, we randomly select 80\% of the category labels as the seen classes, and the remaining 20\%  as the unseen classes. The seen category labels are used for training the model; during testing, the names of unseen categories are used as queries for retrieving instances from the unseen categories. Since there is no attribute information for these three multi-modal datasets, we use Word2Vec \cite{Mikolov2013Distributed}
with skipgram and negative sampling to assign each category with a 500-dimension attribute vector. %{\color{red}[How about the size of the attribute vector?]}. %{\color{red}[Some ZSH also do so, so we suffers the same issue of them???]}

\vspace{-1pt}
\textbf{Comparing methods}: Seven related and representative  methods are adopted for comparison.
\vspace{-2.5pt}
\begin{itemize}
    \item CMSSH (Cross-modal Similarity Sensitive Hashing) \cite{Bronstein2010Data} treats hash code learning as a binary classification problem, and  learns the hash functions using a boosting method.
    \item SCM (Semantic Correlation Maximization) \cite{Zhang2014Large} optimizes the hashing functions by maximizing the correlation between two modalities with respect to semantic labels; it includes two versions, SCM-orth and SCM-seq.  SCM-orth learns hash functions by direct eigen-decomposition with orthogonal constraints for balancing coding functions, and SCM-seq can learn more efficiently hash functions in a sequential manner without the orthogonal constraints.
    \item QCH (Quantized Correlation Hashing)   \cite{Wu2015Quantized}  simultaneously optimizes the quality loss over all data modalities and the pairwise relationships between modalities.
    \item SePH (Semantics Preserving Hashing)   \cite{Lin2017Cross} is a probability-based hashing method, which generates one unified hash code for all observed views by considering the semantic consistency between views.
    \item  DCMH (Deep Cross-modal Hashing) \cite{Jiang2017Deep} is a deep learning based solution that uses the semantic similarity between instances for cross-modal hashing.
    \item RDCMH (Ranking-based deep cross-modal hashing) \cite{liu2019ranking} seamlessly integrates
deep feature learning with semantic ranking based hashing; it can  preserve multi-level semantic similarity between multi-label objects for cross-modal hashing.
    \item AgNet (Attribute-Guided Network for Cross-Modal Zero-Shot Hashing) \cite{Zhong2018Attribute} aligns different modality data into a semantically enriched attribute space and  uses the attribute vectors to guide the generation of hash codes for image and text within the same network.
\end{itemize}
 The first five methods are classical cross-modal hashing methods that use only labeled instances and hand-crafted features.  DCMH, RDCMH, and AgNet additionally use deep learning techniques to learn nonlinear features. Unlike other comparing methods, only CZHash and RDCMH can make use of labeled and unlabeled instances; only CZHash and AgNet take into account the unseen category labels when learning the hashing codes. We adopted the available code for the comparing  methods, and tuned the parameters based on the suggestions given in the respective code or papers. As for CZHash, the regularization parameters $\alpha$ and $\beta$ in Eq. (\ref{Eq7}) are set to 1 and 1, respectively. The number of iterations to optimize Eq. (\ref{Eq7}) is fixed to 500. Further parameter analysis of CZHash is provided in Subsection \ref{parm}. All the experiments are conducted on a server with  Intel E5-2650v3, 256GB RAM  and Ubuntu 16.04.01 OS. The code of CZHash will be public later.
% Binary embedding based Zero-Shot Learning(BZSL)\cite{shen2019scalable} jointly learn two binary coding functions  to map class embeddings and visual features, respectively, into the binary embedding space which could well preserve the discriminative category. (vii)

\textbf{Evaluation metric:}
We adopt the widely used Mean Average Precision (MAP) \cite{Bronstein2010Data,Lin2017Cross,Zhang2014Large} to measure the retrieval performance of all cross-view hashing methods. The formal definition of MAP is  as follows:
\begin{equation}
MAP=\frac{1}{|\mathcal{Q}|}\sum\limits_{i=1}^{|\mathcal{Q}|}\frac{1}{t_i} {\sum\limits_{r=1}^{t_i}{P(r)\delta(r)}}
\end{equation}
where $\mathcal{Q}$ is the query set with  size equal to $|\mathcal{Q}|$. For the $i$-th query, $\frac{1}{t_i} {\sum\limits_{r=1}^{t_i}{P(r)\delta(r)}} $ denotes the average precision (AP), $t_i$ is the number of ground-truth relevant instances in the retrieval set, and $P(r)$ denotes the precision of the top $r$ retrieved results. $\delta(r)=1$ if the $r$-th retrieved result is a true neighbor of the query, otherwise $\delta(r)=0$. A larger MAP value corresponds to a better retrieval performance.

%{\color{red}Precision within a radius should be used for experiment. [Consider about that if we can make it before the due date.]}

\subsection{Results under different scenarios}
\label{Scena}
To  study thoroughly the performance of CZHash and of the comparing methods, we conduct four different experiments:
(\textbf{a}) cross-modal hashing baseline experiments; (\textbf{b}) cross-modal zero-shot hashing experiments; (\textbf{c}) semi-supervised cross-modal zero-shot experiments; (\textbf{d}) semi-supervised cross-modal zero-shot hashing experiments with different label spaces for different modalities. For (\textbf{a}), we use  80\% of the instances for training, and the rest for testing. For (\textbf{b}), based on (\textbf{a}), we randomly select 80\% of the categories as the seen labels, and utilize the instances with seen category labels for training, and the remaining instances with unseen category labels for testing. For (\textbf{c}), based on (\textbf{b}),  we randomly mask the labels of 70\% of the training instances, and use them as unlabeled training instances. For (\textbf{d}), based on (\textbf{c}), we randomly select different 80\% of labels in the two label spaces as the seen category labels for the image and text modalities, respectively. We set the hashing code length to 16, 32, 64, and 128 bits to study the performance under different code lengths, and repeat the experiments 10 times to avoid the random effect of data partition.   The variance across ten independent runs is always smaller than 0.2\%, so for brevity we just report the average MAP. The results are reported in Tables \ref{Table3}-\ref{Table6},  and the best results under each configuration are \textbf{boldfaced}. Given the page limit and the performance margin been between deep learning based hashing and non-deep hashing methods, we only report the results of deep-based methods (DCMH, RDCMH, AgNet and DRCMH) in Tables \ref{Table5}-\ref{Table6} under scenario (\textbf{c}) and (\textbf{d}), respectively. We observe that all the comparing methods cannot be directly applied to the scenario (\textbf{d}). For comparison, we use the common labels of the two modalities to train the comparing methods. The results of AgNet on Nus-wide are not reported since its runtime is too long (more than 4 days for one round) and the results can not complete before the deadline.

\begin{table*}[h!tbp]
\label{Table3}
\centering
\scriptsize
\caption{
Results (MAP) on three  datasets with complete data.}
\vspace{-1em}
\begin{tabular}{p{0.8cm}|p{1.5cm}|p{0.78cm}|p{0.78cm}|p{0.78cm}|p{0.78cm}||p{0.78cm}|p{0.78cm}|p{0.78cm}|p{0.78cm}||p{0.78cm}|p{0.78cm}|p{0.78cm}|p{0.78cm}}
\hline
 &  & \multicolumn{4}{c}{\textbf{Mirflickr}} & \multicolumn{4}{c}{\textbf{Nus-wide}} & \multicolumn{4}{c}{\textbf{Wiki}}\\\hline
 &Methods &16bits  &32bits &64bits &128bits &16bits  &32bits &64bits &128bits  &16bits  &32bits &64bits &128bits	\\\hline
 \multirow{9}{*}{\tabincell{c}{Image\\ vs.\\ Text}}
&CMSSH &	$0.5616$ &	$0.5555$ &	$0.5513$ &	$0.5484$   &	 $0.3414$ &	$0.3336$ &	$0.3282$ &	$0.3261$ & $0.1694$ &	$0.1523$ &	$0.1447$ &	$0.1434$   \\
&SCM-seq &$0.5721$ &	$0.5607$ &	$0.5535$ &	$0.5482$ &	$0.3623$ &	$0.3646$ &	$0.3703$ &	$0.3721$ & $0.1577$ &	$0.1434$ &	$0.1376$ &	$0.1358$   \\
&SCM-orth &	$0.6041$ &	$0.6112$ &	$0.6176$ &	$0.6232$  &	$0.4651$ &	$0.4714$ &	$0.4822$ &	$0.4851$ & $0.2341$ &	$0.2411$ &	$0.2443$ &	$0.2564$   \\
&QCH &	$0.6232$ &	$0.6256$ &	$0.6268$ &	$0.6293$   & $0.4752$ &	$0.4793$ &	$0.4812$ &	$0.4866$ & $0.2578$ &	$0.2591$ &	$0.2603$ &	$0.2612$ \\
&SePH &	$0.6573$ &	$0.6603$ &	$0.6616$ &	$0.6637$   & $0.4787$ &	$0.4869$ &	$0.4888$ &	$0.4932$ & $0.2836$ &	$0.2859$ &	$0.2879$ &	$0.2863$  \\
&DCMH &$0.7411$ &	$0.7465$ &	$0.7485$ &	$0.7493$ & $0.5903$ &	$0.6031$ &	$0.6093$ &	$0.6124$  &  $0.2673$ &	$0.2684$ &	$0.2687$ &	$0.2748$ \\
&RDCMH &$0.7723$ &	$0.7735$ &	$0.7789$ &	$0.7810$ & $0.6231$ &	$0.6236$ &	$0.6273$ &	$0.6302$  &  $0.2943$ &	$0.2968$ &	$0.3001$ &	$0.3042$ \\
&AgNet &	$0.6431 $ &	$0.6442 $ &	$0.6455 $ &	$0.6471 $  &	 $- $ &	$- $ &	$- $ &	$- $ & $0.2667$ &	$0.2674$ &	$0.2711$ &	$0.2734$   \\

&CZHash &	${\mathbf{0.7812 }}$ &	${\mathbf{0.7831
 }}$ &		${\mathbf{0.7862 }}$ & ${\mathbf{0.7874 }}$&	${\mathbf{0.6411 }}$ &		${\mathbf{0.6434 }}$ &	${\mathbf{0.6457 }}$ &${\mathbf{0.6468 }}$ & ${\mathbf{0.2998 }}$ &		${\mathbf{0.3017 }}$ &	${\mathbf{0.3035 }}$ &${\mathbf{0.3063 }}$\\
\hline

  \multirow{9}{*}{\tabincell{c}{Text \\ vs. \\ Image}}
&CMSSH &	 $0.5616$ &	$0.5551$ &	$0.5506$ &	$0.5475$  &	 $0.3392$ &	$0.3321$ &	$0.3272$ &	$0.3256$ &  $0.1578$ &	$0.1384$ &	$0.1331$ &	$0.1256$ \\
&SCM-seq &	 $0.5694$ &	$0.5611$ &	$0.5544$ &	$0.5497$  &	$0.3412$ &	$0.3459$ &	$0.3472$ &	$0.3539$ & $0.1521$ & $0.1561$ &	$0.1371$ &	$0.1261$  \\
&SCM-orth &	$0.6055$ &	$0.6154$ &	$0.6238$ &	$0.6299$    &	$0.4370$ &	$0.4428$ &	$0.4504$ &	$0.2235$  & $0.2257$ &	$0.2459$ &	$0.2482$ &	$0.2518$ \\
&QCH &	$0.6205$ &	$0.6237$ &	$0.6259$ &	$0.6286$   & $0.4349$ &	$0.4387$ &	$0.4412$ &	$0.4425$ & $0.2872$ &	$0.2891$ &	$0.2907$ &	$0.2923$ \\
&SePH & $0.6481$ &	$0.6521$ &	$0.6545$ &	$0.6534$    & $0.4489$ &	$0.4539$ &	$0.4587$ &	$0.4621$ &  ${\mathbf{0.5345}}$ &	${\mathbf{0.5351}}$ &	${\mathbf{0.5471}}$ &	${\mathbf{0.5506}}$ \\
&DCMH & $0.7827$ &	$0.7901$ &	$0.7932$ &	$0.7956$   & $0.6389$ &	$0.6511$ &	$0.6571$ &	$0.6589$ & $0.2712$ &	$0.2751$ &	$0.2812$ &	$0.2789$  \\
&RDCMH & $0.7931$ &	$0.7924$ &	$0.8001$ &	$0.8024$   & $0.6641$ &	$0.6685$ &	$0.6694$ &	$0.6703$ & $0.2931$ &	$0.2956$ &	$0.3012$ &	$0.3035$  \\
&AgNet &	 $0.6401 $ &	$0.6413 $ &	$0.6426 $ &	$0.6442 $   &	$ -$ &	$- $ &	$- $ &	$- $ & $0.2643$ &	$0.2661$ &	$0.2672$ &	$0.2693$    \\

&CZHash &	${\mathbf{0.7964 }}$ &		${\mathbf{0.7989 }}$ &	${\mathbf{0.8025 }}$ &${\mathbf{0.8037 }}$&	${\mathbf{0.6755 }}$ &		${\mathbf{0.6763 }}$ &	${\mathbf{0.6789 }}$ &${\mathbf{0.6796 }} $&	${{0.3016 }}$ &		${{0.3025}}$ &	${{0.3044 }}$ &${{0.3061 }}$\\\hline
\end{tabular}
\label{Table3}
\end{table*}

\begin{table*}[h!tbp]
\label{Table4}
\centering
\scriptsize
\caption{
Results (MAP) on three  datasets with zero-shot data.}
\vspace{-1em}
\begin{tabular}{p{0.8cm}|p{1.5cm}|p{0.78cm}|p{0.78cm}|p{0.78cm}|p{0.78cm}||p{0.78cm}|p{0.78cm}|p{0.78cm}|p{0.78cm}||p{0.78cm}|p{0.78cm}|p{0.78cm}|p{0.78cm}}
\hline
 &  & \multicolumn{4}{c}{\textbf{Mirflickr}} & \multicolumn{4}{c}{\textbf{Nus-wide}} & \multicolumn{4}{c}{\textbf{Wiki}}\\\hline
 &Methods &16bits  &32bits &64bits &128bits &16bits  &32bits &64bits &128bits  &16bits  &32bits &64bits &128bits	\\\hline
 \multirow{9}{*}{\tabincell{c}{Image\\ vs.\\ Text}}
&CMSSH &	$0.5175$ &	$0.5111$ &	$0.5073$ &	$0.5043$ &	 $0.3152$ &	$0.3081$ &	$0.3015$ &	$0.2973$ & $0.1403$ &	$0.1376$ &	$0.1341$ &	$0.1302$   \\
&SCM-seq & $0.5325$ &	$0.5213$ &	$0.5141$ &	$0.5097$ &	$0.3342$ &	$0.3315$ &	$0.3287$ &	$0.3254$ & $0.1489$ &	$0.1453$ &	$0.1431$ &	$0.1406$   \\
&SCM-orth &	$0.5753$ &	$0.5862$ &	$0.5961$ &	$0.6068$ &	$0.4437$ &	$0.4415$ &	$0.4463$ &	$0.4491$ & $0.2102$ &	$0.2126$ &	$0.2164$ &	$0.2183$   \\
&QCH &	$0.5814$ &	$0.5835$ &	$0.5869$ &	$0.5913$ & $0.4468$ &	$0.4495$ &	$0.4512$ &	$0.4538$ & $0.2213$ &	$0.2245$ &	$0.2281$ &	$0.2308$ \\
&SePH &	$0.6013$ &	$0.6059$ &	$0.6103$ &	$0.6145$ & $0.4493$ &	$0.4524$ &	$0.4567$ &	$0.4606$ & $0.2412$ &	$0.2453$ &	$0.2489$ &	$0.2507$  \\
&DCMH &$0.6814$ &	$0.6845$ &	$0.6873$ &	$0.6926$ &$0.5321 $ &	$0.5367 $ &	$0.5392 $ &	$0.5436 $  &  $0.2212 $ &	$0.2237 $ &	$0.2268 $ &	$0.2285 $ \\
&RDCMH &$0.6879$ &	$0.6894$ &	$0.6923$ &	$0.6939$ &$0.5394 $ &	$0.5421 $ &	$0.5446 $ &	$0.5477 $  &  $0.2263 $ &	$0.2291 $ &	$0.2308 $ &	$0.2331 $ \\
&AgNet &	$0.6274 $ &	$0.6281 $ &	$0.6299 $ &	$0.6337 $  &	 $- $ &	$- $ &	$- $ &	$- $ & $0.2313$ &	$0.2326$ &	$0.2337$ &	$0.2345$    \\

%&  $\textbf{DRCMH-NW}$ &	${\mathbf{0.1062}}$ &	${\mathbf{0.1064}}$  &	${\mathbf{0.1067}}$ &	${\mathbf{0.1058}}$\\
&CZHash &	${\mathbf{0.7302 }}$ &	${\mathbf{0.7334
 }}$ &		${\mathbf{0.7351 }}$ & ${\mathbf{0.7363 }}$&	${\mathbf{0.5982 }}$ &		${\mathbf{0.6017 }}$ &	${\mathbf{0.6033 }}$ &${\mathbf{0.6059 }}$ & ${\mathbf{0.2543 }}$ &		${\mathbf{0.2551 }}$ &	${\mathbf{0.2576 }}$ &${\mathbf{0.2581 }}$\\
\hline

  \multirow{9}{*}{\tabincell{c}{Text \\ vs. \\ Image}}
&CMSSH &	$0.5114$ &	$0.5073$ &	$0.5053$ &	$0.5031$ &	 $0.3006$ &	$0.2984$ &	$0.2955$ &	$0.2918$ &  $0.1342$ &	$0.1315$ &	$0.1294$ &	$0.1256$ \\
&SCM-seq &	 $0.5226$ &	$0.5089$ &	$0.5032$ &	$0.5025$ &	$0.3315$ &	$0.3087$ &	$0.3042$ &	$0.3021$ & $0.1402$ & $0.1384$ &	$0.1352$ &	$0.1332$  \\
&SCM-orth &	$0.5475$ &	$0.5522$ &	$0.5601$ & $0.5632$ &	$0.4106$ &	$0.4135$ &	$0.4184$ &	$0.4205$  & $0.2012$ &	$0.2035$ &	$0.2051$ &	$0.2072$ \\
&QCH &	$0.5641$ &	$0.5683$ &	$0.5712$ &	$0.5735$ & $0.4144$ &	$0.4162$ &	$0.4175$ &	$0.4180$ & $0.2423$ &	$0.2448$ &	$0.2475$ &	$0.2483$ \\
&SePH & $0.5803$ &	$0.5824$ &	$0.5869$ &	$0.5964$ & $0.4215$ &	$0.4238$ &	$0.4293$ &	$0.4312$ &  ${\mathbf{0.4975}}$ &	${\mathbf{0.4996}}$ &	${\mathbf{0.5013}}$ &	${\mathbf{0.5044}}$ \\
&DCMH & $0.6542$ &	$0.6578$ &	$0.6613$ &	$0.6675$ & $0.5935 $ &	$0.5966 $ &	$0.6012 $ &	$0.6031 $ & $0.2425 $ &	$0.2448 $ &	$0.2466 $ &	$0.2483 $  \\
&RDCMH & $0.6601$ &	$0.6623$ &	$0.6649$ &	$0.6683$ & $0.5987 $ &	$0.6014 $ &	$0.6037 $ &	$0.6065 $ & $0.2465 $ &	$0.2483 $ &	$0.2497 $ &	$0.2502 $  \\
&AgNet &	 $0.6255 $ &	$0.6264 $ &	$0.6273 $ &	$0.6288 $   &	$ -$ &	$- $ &	$- $ &	$- $ &  $0.2462$ &	$0.2471$ &	$0.2487$ &	$0.2511$    \\

&CZHash &	${\mathbf{0.7092 }}$ &		${\mathbf{0.7113
 }}$ &	${\mathbf{0.7132 }}$ &${\mathbf{0.7148 }}$&	${\mathbf{0.6286 }}$ &		${\mathbf{0.6297}}$ &	${\mathbf{0.6325 }}$ &${\mathbf{0.6339 }} $&	${{0.2526 }}$ &		${0.2541}$ &	${{0.2563 }}$ &${{0.2587}}$\\\hline
\end{tabular}
\label{Table4}
\end{table*}

\begin{table*}[h!tbp]
\label{Table5}
\centering
\scriptsize
\caption{
Results (MAP) on three  datasets with semi-supervised zero-shot data. }
\vspace{-1em}
%{\color{red}[Zhong Ji shows better results than DCMH, we did not, there is a doubt.]}{\color{blue} AgNet is lower than DCMH in mir and nus, but higher in Wiki, this is reasonable beacues AgNet may be not adaptive on multi-label datasets.}}
\begin{tabular}{p{0.8cm}|p{1.5cm}|p{0.78cm}|p{0.78cm}|p{0.78cm}|p{0.78cm}||p{0.78cm}|p{0.78cm}|p{0.78cm}|p{0.78cm}||p{0.78cm}|p{0.78cm}|p{0.78cm}|p{0.78cm}}
\hline
 &  & \multicolumn{4}{c}{\textbf{Mirflickr}} & \multicolumn{4}{c}{\textbf{Nus-wide}} & \multicolumn{4}{c}{\textbf{Wiki}}\\\hline
 &Methods &16bits  &32bits &64bits &128bits &16bits  &32bits &64bits &128bits  &16bits  &32bits &64bits &128bits	\\\hline
 \multirow{5}{*}{\tabincell{c}{Image\\ vs.\\ Text}}
%&CMSSH &$0.1576$ &	$0.1554$ &	$0.1542$ &	$0.1532$   &	$0.0792$ &	$0.0786$ &	$0.0775$ &	$0.0769$  & $0.0174$ &	$0.0172$ &	$0.0168$ &	$0.0167$    \\
%&SCM-seq &$0.1599$ &	$0.1575$ &	$0.1545$ &	$0.1533$  &	$0.0811$ &	$0.0796$ &	$0.0783$ &	$0.0778$ & $0.0196$ &	$0.0193$ &	$0.0192$ &	$0.0187$  \\
%&SCM-orth &	$0.1569$ &	$0.1573$ &	$0.1584$ &	$0.1605$   &$0.1012$ &	$0.1026$ &	$0.1033$ &	$0.1041$ & $0.0204$ &	$0.0211$ &	$0.0215$ &	$0.0223$  \\
%&QCH &	$0.1651$ &	$0.1667$ &	$0.1683$ &	$0.1715$  &	 $0.1053$ &	$0.1067$ &	$0.1079$ &	$0.1083$ & $0.0277$ &	$0.0284$ &	$0.0289$ &	$0.0291$  \\
%&SePH &	$0.1847$ &	$0.1866$ &	$0.1879$ &	$0.1892$   &$0.1242$ &	$0.1257$ &	$0.1262$ &	$0.1281$  &$0.0615$ &	$0.0619$ &	$0.0622$ &	$0.0624$   \\
&DCMH &$0.1695$ &	$0.1712$ &	$0.1733$ &	$0.1739$ &$0.1153$ &	$0.1167$ &	$0.1174$ &	$0.1188$ & $0.0542$ &	$0.0547$ &	$0.0556$ &	$0.0559$   \\
&RDCMH &$0.1915$ &	$0.1937$ &	$0.1963$ &	$0.1975$ &$0.1248$ &	$0.1269$ &	$0.1283$ &	$0.1301$ & $0.0983$ &	$0.1014$ &	$0.1026$ &	$0.1041$   \\
&AgNet &	$ 0.1613 $ &	$0.1633 $ &	$0.1642 $ &	$0.1656 $  &	 $- $ &	$- $ &	$- $ &	$- $ & $0.0673$ &	$0.0689$ &	$0.0694$ &	$0.0711$   \\

%&  $\textbf{DRCMH-NW}$ &	${\mathbf{0.1062}}$ &	${\mathbf{0.1064}}$  &	${\mathbf{0.1067}}$ &	${\mathbf{0.1058}}$\\
&CZHash &	${\mathbf{0.2126}}$ &	${\mathbf{0.2135}}$ &		${\mathbf{0.2141}}$ & ${\mathbf{0.2147
}}$&	${\mathbf{0.1733}}$ &		${\mathbf{0.1756}}$ &	${\mathbf{0.1771
 }}$ &${\mathbf{0.1783
 }}$ & ${\mathbf{0.1214
 }}$ &		${\mathbf{0.1233
 }}$ &	${\mathbf{0.1247
 }}$ &${\mathbf{0.1251
 }}$\\
\hline

  \multirow{5}{*}{\tabincell{c}{Text \\ vs. \\ Image}}
%&CMSSH &$0.1551$ &	$0.1541$ &	$0.1536$ &	$0.1533$   &$0.0743$ &	$0.0736$ &	$0.0724$ &	$0.0715$ & $0.0154$ &	$0.0151$ &	$0.0149$ &	$0.0144$ \\
%&SCM-seq &$0.1572$ &	$0.1546$ &	$0.1521$ &	$0.1517$   &$0.0782$ &	$0.0769$ &	$0.0753$ &	$0.0744$ &
% $0.0177$ &	$0.0174$ &	$0.0171$ &	$0.0166$  \\
%&SCM-orth &	$0.1548$ &	$0.1553$ &	$0.1563$ &	$0.1564$   &$0.0986$ &	$0.1007$ &	$0.1018$ &	$0.1023$ &$0.0198$ &	$0.0203$ &	$0.0207$ &	$0.0211$  \\
%&QCH &	$0.1613$ &	$0.1622$ &	$0.1649$ &	$0.1663$  &	 $0.1022$ &	$0.1035$ &	$0.1044$ &	$0.1062$ & $0.0246$ &	$0.0251$ &	$0.0253$ &	$0.0262$  \\
%&SePH & $0.1793$ &	$0.1814$ &	$0.1836$ &	$0.1851$   &	 $0.1197$ &	$0.1226$ &	$0.1232$ &	$0.1249$ & $0.0583$ &	$0.0589$ &	$0.0594$ &	$0.0607$  \\
&DCMH & $0.1657$ &	$0.1664$ &	$0.1683$ &	$0.1702$ &	 $0.1106$ &	$0.1123$ &	$0.1146$ &	$0.1169$ &
 $0.0512$ &	$0.0517$ &	$0.0523$ &	$0.0526$  \\
 &RDCMH & $0.1874$ &	$0.1886$ &	$0.1895$ &	$0.1902$ &	 $0.1253$ &	$0.1266$ &	$0.1272$ &	$0.1281$ &
 $0.0924$ &	$0.0931$ &	$0.0945$ &	$0.0952$  \\
&AgNet &	 $0.1623 $ &	$0.1631 $ &	$0.1635 $ &	$0.1652 $   &	$ -$ &	$- $ &	$- $ &	$- $ &  $0.0626$ &	$0.0637$ &	$0.0648$ &	$0.0656$   \\

&CZHash &	${\mathbf{0.2016 }}$ &		${\mathbf{0.2023}}$ &${\mathbf{0.2027}}$ &${\mathbf{0.2031}}$&	${\mathbf{0.1721
}}$ &		${\mathbf{0.1736}}$ &	${\mathbf{0.1743}}$ &${\mathbf{0.1748}} $&	${\mathbf{0.1178}}$ &	${\mathbf{0.1196}}$ &	${\mathbf{0.1218}}$ &${\mathbf{0.1232}}$\\\hline
\end{tabular}
\label{Table5}
\end{table*}

\begin{table*}[h!tbp]
\label{Table6}
\centering
\scriptsize
\caption{
Results (MAP) on three  datasets with semi-supervised zero-shot data and different label spaces. }
\vspace{-1em}
\begin{tabular}{p{0.8cm}|p{1.5cm}|p{0.78cm}|p{0.78cm}|p{0.78cm}|p{0.78cm}||p{0.78cm}|p{0.78cm}|p{0.78cm}|p{0.78cm}||p{0.78cm}|p{0.78cm}|p{0.78cm}|p{0.78cm}}
\hline
 &  & \multicolumn{4}{c}{\textbf{Mirflickr}} & \multicolumn{4}{c}{\textbf{Nus-wide}} & \multicolumn{4}{c}{\textbf{Wiki}}\\\hline
 &Methods &16bits  &32bits &64bits &128bits &16bits  &32bits &64bits &128bits  &16bits  &32bits &64bits &128bits	\\\hline
 \multirow{4}{*}{\tabincell{c}{Image\\ vs.\\ Text}}
%&CMSSH &	$0.1131$ &	$0.1124$ &	$0.1121$ &	$0.1118$   &	$0.0536$ &	$0.0531$ &	$0.0527$ &	$0.0524$ &
% $0.0118$ &	$0.0116$ &	$0.0115$ &	$0.0113$    \\
%&SCM-seq &$0.1164$ &	$0.1145$ &	$0.1138$ &	$0.1151$ &	$0.0572$ &	$0.0569$ &	$0.0561$ &	$0.0559$ &
%$0.0119$ &	$0.0117$ &	$0.0116$ &	$0.0114$  \\
%&SCM-orth &	$0.1165$ &	$0.1172$ &	$0.1175$ &	$0.1183$ &	$0.0617$ &	$0.0623$ &	$0.0626$ &	$0.0631$ &
%$0.0167$ &	$0.0169$ &	$0.0172$ &	$0.0173$ \\
%&QCH &	$0.1171$ &	$0.1178$ &	$0.1183$ &	$0.1192$ &
%$0.0621$ &	$0.0627$ &	$0.0633$ &	$0.0635$ &
%$0.0193$ &	$0.0197$ &	$0.0203$ &	$0.0209$  \\
%&SePH &	$0.1231$ &	$0.1241$ &	$0.1255$ &	$0.1263$ &
%$0.0897$ &	$0.0916$ &	$0.0921$ &	$0.0933$ &
%$0.0415$ &	$0.0424$ &	$0.0431$ &	$0.0437$   \\
&DCMH & $0.1256$ &	$0.1262$ &	$0.1271$ &	$0.1283$ &
$0.0852$ &	$0.0857$ &	$0.0864$ &	$0.0873$ &
$0.0427$ &	$0.0439$ &	$0.0451$ &	$0.0458$   \\
&RDCMH &$0.1511$ &	$0.1519$ &	$0.1526$ &	$0.1533$ &
	$0.1036$ &	$0.1041$ &	$0.1047$ &	$0.1052$ &
$0.0541$ &	$0.0549$ &	$0.0554$ &	$0.0562$ \\
&AgNet &	$0.1231 $ &	$0.1239 $ &	$0.1245 $ &	$0.1268 $  &	 $- $ &	$- $ &	$- $ &	$- $ & $0.0436$ &	$0.0445$ &	$0.0454$ &	$0.0462$   \\

%&  $\textbf{DRCMH-NW}$ &	${\mathbf{0.1062}}$ &	${\mathbf{0.1064}}$  &	${\mathbf{0.1067}}$ &	${\mathbf{0.1058}}$\\
&CZHash &	${\mathbf{0.1873 }}$ &	${\mathbf{0.1899 }}$ &		${\mathbf{0.1917 }}$ & ${\mathbf{0.1926 }}$&	${\mathbf{0.1481
 }}$ &		${\mathbf{0.1492}}$ &	${\mathbf{0.1511}}$ &${\mathbf{0.1519
 }}$ & 	${\mathbf{0.0982}}$ &		${\mathbf{0.0997}}$ &	${\mathbf{0.1012}}$ &${\mathbf{0.1019 }}$\\
\hline

  \multirow{4}{*}{\tabincell{c}{Text \\ vs. \\ Image}}
%&CMSSH &$0.1126$ &	$0.1123$ &	$0.1121$ &	$0.1121$ &
%	$0.0526$ &	$0.0524$ &	$0.0521$ &	$0.0517$ &
%$0.0114$ &	$0.0113$ &	$0.0111$ &	$0.011$ \\
%&SCM-seq &$0.1141$ &	$0.1135$ &	$0.1134$ &	$0.1131$ & $0.0563$ &	$0.0559$ &	$0.0556$ &	$0.0554$ &
% $0.0116$ &	$0.0114$ &	$0.0113$ &	$0.0111$ \\
%&SCM-orth &$0.1132$ &	$0.1136$ &	$0.1145$ &	$0.1148$ &	$0.0604$ &	$0.0609$ &	$0.0612$ &	$0.0617$ &
% $0.0166$ &	$0.0169$ &	$0.0171$ &	$0.0171$  \\
%&QCH &	$0.1126$ &	$0.1135$ &	$0.1139$ &	$0.1141$ &
%$0.0616$ &	$0.0619$ &	$0.0625$ &	$0.0627$ &
% $0.0187$ &	$0.0196$ &	$0.0201$ &	$0.0204$  \\
%&SePH & $0.1169$ &	$0.1187$ &	$0.1193$ &	$0.1204$ &
%$0.0887$ &	$0.0894$ &	$0.0911$ &	$0.0926$ &
%$0.0409$ &	$0.0416$ &	$0.0425$ &	$0.0432$  \\
&DCMH &$0.1189$ &	$0.1195$ &	$0.1203$ &	$0.1208$ &
$0.0831$ &	$0.0839$ &	$0.0841$ &	$0.0848$ &
$0.0415$ &	$0.0423$ &	$0.0429$ &	$0.0431$   \\
&RDCMH &$0.1418$ &	$0.1432$ &	$0.1437$ &	$0.1455$ &
$0.0986$ &	$0.1004$ &	$0.1015$ &	$0.1019$ &
$0.0531$ &	$0.0536$ &	$0.0540$ &	$0.0547$  \\
&AgNet &	 $0.1176 $ &	$0.1189 $ &	$0.1194 $ &	$0.1201 $   &	$ -$ &	$- $ &	$- $ &	$- $ &   $0.0417$ &	$0.0425$ &	$0.0432$ &	$0.0437$    \\

&CZHash &	${\mathbf{0.1795 }}$ &		${\mathbf{0.1807 }}$ &	${\mathbf{0.1816 }}$ &${\mathbf{0.1822 }}$&	${\mathbf{0.1437}}$ &		${\mathbf{0.1453 }}$ &	${\mathbf{0.1475 }}$ &${\mathbf{0.1498 }} $& ${\mathbf{0.0936}}$ &		${\mathbf{0.0949}}$ &	${\mathbf{0.0971}}$ &${\mathbf{0.0995 }} $\\\hline
\end{tabular}
\label{Table6}
\end{table*}

Given the Tables, we have the following observations.\\
(1) \textbf{deep features vs. handcraft features based methods}:   Deep hashing methods (DCMH, RDCMH, AgNet, and CZHash) consistently achieve a performance which is superior to the other comparing methods in most scenarios. This proves that  features learned through a deep neural network are more suitable for hashing function learning than hand-crafted ones.  Comparing with other deep-based methods,  CZHash  achieves the best performance in most cases. This is because CZHash  not only  jointly models the inter-modal and the intra-modal similarity preserving losses to learn a more faithfully semantic presentation,  but also considers a category space learning loss and  the quantitative loss to learn more adaptive hashing codes. In particular, we also observe from  Table \ref{Table3} that SePH obtains better results for `Text vs. Image' retrieval on Wiki. This is possible because of the adaptability of its probability-based strategy on small datasets.  An unexpected observation is that the performance of CMSSH and SCM-Orth decreases as the length of hash codes increases. This might be caused by the imbalance between bits in the hash codes learned by singular value or eigenvalue decomposition.

(2) \textbf{AgNet and CZHash vs. other non zero-shot methods}: In Table \ref{Table4}, all the non zero-shot methods have lower MAP results than those in Table \ref{Table3}, while AgNet and CZHash still achieve promising results.
This is because these hashing methods (except AgNet and CZHash) are not designed for zero-shot cross-modal learning, so they don't have the ability of recognizing  unseen classes,  where AgNet and CZHash can achieve zero-shot hashing to reduce the impact of unseen class. CZHash still outperforms AgNet, which shows not only the superiority of utilizing the introduced composite similarity to guide deep feature learning learning,  but also the advantages of jointly optimizing deep feature learning, category space learning, and hash-code learning processes.

(3) \textbf{RDCMH and CZHash vs. other supervised methods}: From Table \ref{Table5},  we can see that all the supervised  methods manifest sharply reduced MAP values, while semi-supervised methods (RDCMH and CZHash) are less effected. This is because label integrity has a significant impact on the effectiveness of supervised hashing methods. All the comparing methods ask for sufficient label information to guide the hashing code learning. Unfortunately, the comparing methods disregard unlabeled data, which instead contribute to a more faithful exploration of the data structure, and to reliable cross-modal hashing codes. This observation proves the effectiveness of the introduced  semi-supervised composite similarity in leveraging unlabeled data to boost  hashing code learning.  CZHash also outperforms RDCMH; this is because CZHash learns category spaces to capture  correlations from seen to unseen classes, which can handle zero-shot hashing cases, whereas RDCMH cannot.

(4) \textbf{CZHash vs. the other methods}: From Table \ref{Table6}, we can see that all comparing methods have reduced results compared to those in Table \ref{Table5}, although they have the same portion of seen labels for each modality as in Table \ref{Table5}. That is because the label spaces of different modalities are not the same, these  non-overlapped labels could provide additional guidance for each modality, which are complementary to each other. However, these comparing methods can only use the overlapped labels of two modalities.  In contrast, CZHash accounts for different label spaces of different modalities, it can leverage non-overlapped and overlapped labels to guide the feature learning. As such, it achieves higher MAP results than other comparing methods.

(5) \textbf{ CZHash vs. AgNet}: From the four tables, we can see that CZHash always achieves better results than AgNet, although both of them are cross-modal zero-shot hashing methods. This is because CZHash has the following advantages with respect to AgNet: (i) CZHash introduces a composite similarity measure to utilize both labeled and unlabeled data to guide feature learning;  (ii) CZHash considers the different label spaces of the various modalities to obtain more abundant supervised information, whereas AgNet requires the same label space; (iii) CZHash  simultaneously optimizes  deep feature learning, category space learning, and hash-code learning, which can lead to more compatible hashing codes, while AgNet isolates the optimization of these learning tasks; (iv) AgNet is designed for single-label retrieval tasks, and may not adapt well to multi-label datasets (Miflickr and Nus-wide) due to the difficulty of obtaining multi-label instance-level attributes. For the last two reasons, AgNet sometimes even loses to handcraft features based comparing methods. In summary, our CZHash is more effective and adaptive for cross-modal zero-shot retrieval tasks.

%{\color{red}[We should highlight the different label spaces and go deeper comparison and analysis!] [Think about it.]}

\subsection{Parameter sensitivity analysis}
\label{parm}
We further explore the sensitivity of the scalar parameters $\alpha$ and  $\beta$ in Eq. (\ref{Eq7}), and report the results on three datasets in Fig. \ref{Fig2}, where the code length is fixed to 16 bits.  We can see that CZHash is slightly sensitive to $\alpha$ and $\beta$ with $\alpha \in [10^{-2}, 10^{2}] $ and $\beta \in [10^{-2}, 10^{2}] $, and achieves the best performance when $\alpha = 1$ and $\beta = 1$. Over-weighting or under-weighting  category space learning and the quantitative loss has a negative impact on the performance. In practice, CZHash under other code lengths (32 bits, 64 bits, 128 bits) also gives similar observations. In summary, effective $\alpha$ and $\beta$ values can be easily selected for CZHash.
\begin{figure}[h!tbp]
\centering
{\label{Fig2}\includegraphics[width=9cm,height=3.3cm]{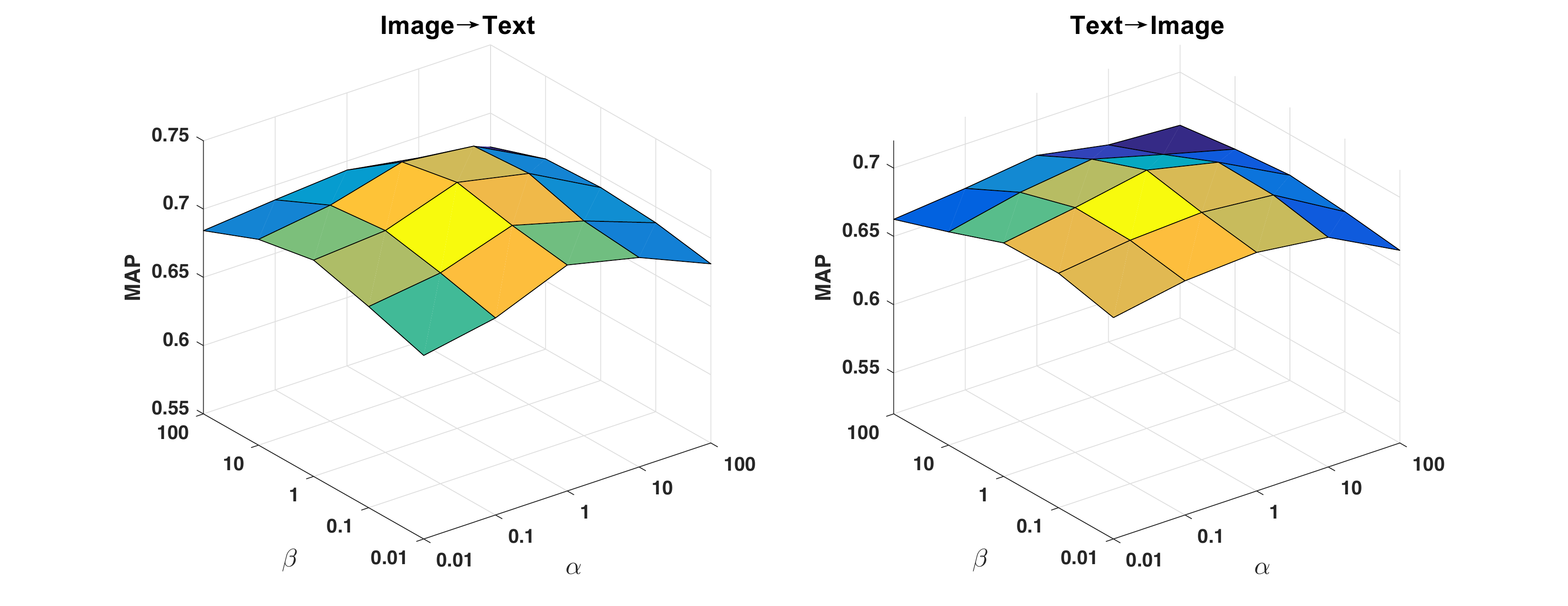}}
\vspace{-6pt}
\caption{MAP vs. $\alpha$ and $\beta$ on Mirfilcker dataset.}
\label{Fig2}
\vspace{-10pt}
\end{figure}

\subsection{Further analysis}
To further investigate the contribution components of CZHash, we introduce  three variants of CZHash, namely CZHash(nFS), CZHash-nLS and CZHash-nJ. CZHash-nFS doesn't adopt the proposed composite similarity measure, but only uses  semantic label similarity. Similarily, CZHash-nLS only utilize the feature similarity without label information.
CZHash-nJ denotes the variant that isolates the optimization of deep feature learning, category space learning, and hashing code function learning, without jointly optimizing them.  Fig. \ref{Fig3} shows the results of these variants on the Mirflickr dataset with semi-supervised zero-shot setting (scenario \textbf{c}) (in Subsection \ref{Scena}). The results on the other datasets show similar results, and are not reported  for space limitation.
\vspace{-4pt}
\begin{figure}[h!tbp]
\centering
{\label{Fig3}\includegraphics[width=9cm,height=3.5cm]{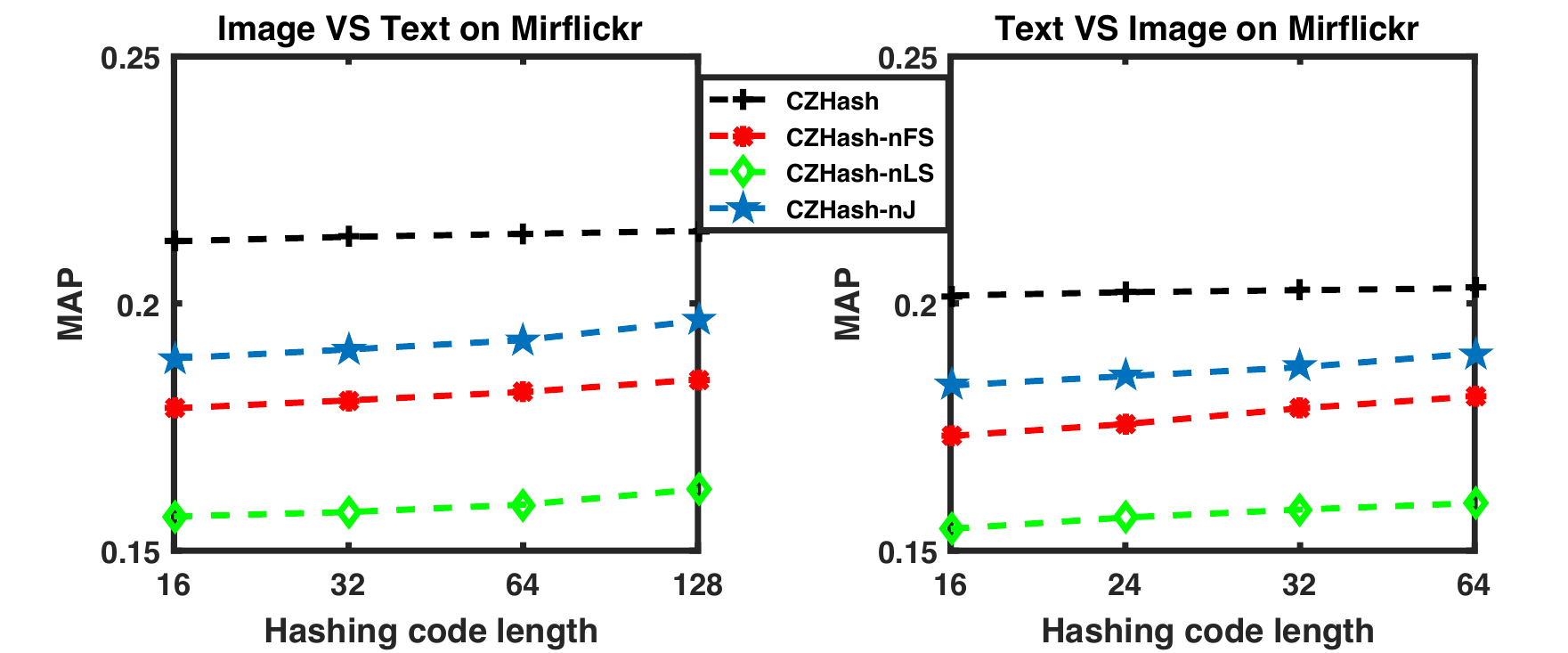}}
\caption{Results of different variants of CZHash on Mirfilcker.}
\label{Fig3}
\end{figure}

From Fig. \ref{Fig3} we can see that CZHash outperforms both CZHash-nFS and CZHash-nLS, which indicates that  the proposed composite similarity measure can handle semi-supervised scenarios, and improve the cross-modal retrieval quality, since it utilizes both labeled and unlabeled data to guide the cross-modal hashing learning. In addition, CZHash achieves a higher accuracy than CZHash-nJ. These results show the advantages of simultaneously optimizing deep feature learning, category space learning, and hash-code learning. %{\color{red}[We lack a variant of CZHash(nLs) only using the feature similarity. [Fix it.]]}
\vspace{-2pt}
\subsection{Convergence curves and runtime analysis}

\begin{figure}[h!tbp]
\centering
{\label{Fig4}\includegraphics[width=9cm,height=2.7cm]{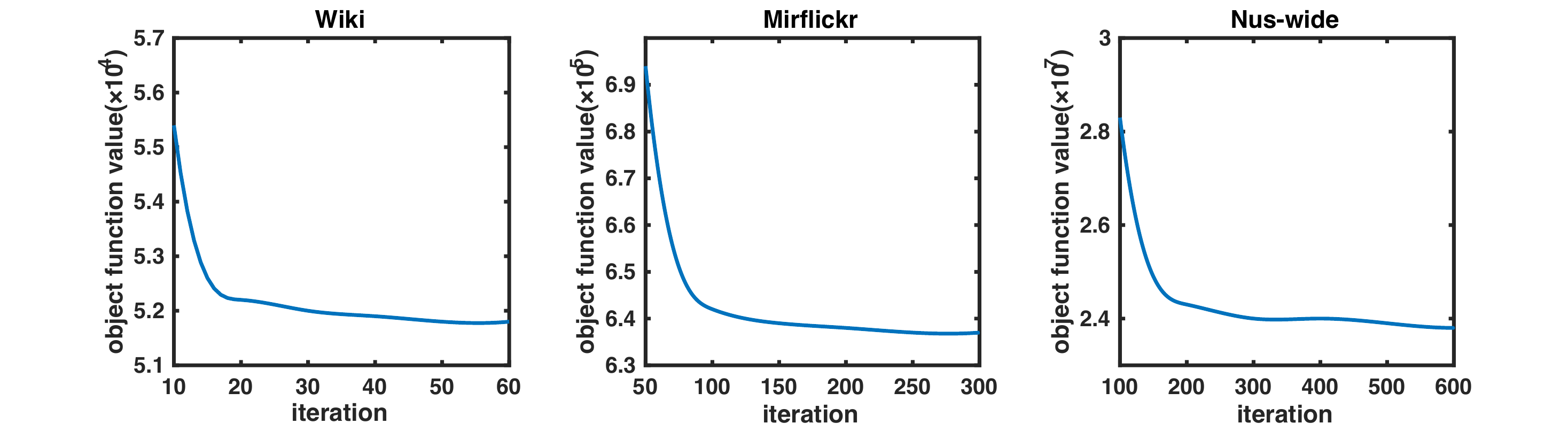}}
\vspace{-15pt}
\caption{Convergence curve of CZHash on three datasets.}
\label{Fig4}
\vspace{-10pt}
\end{figure}

We further plot the convergence curves of CZHash on three datasets in Fig. \ref{Fig4}. CZHash reaches a stable objective function value at the 20-th, 100-th and 200-th epoch on Wiki, Mirflickr and Nus-wide, respectively. And the runtime of CZHash and AgNet is statistically  shown in Table \ref{Table8}, we can see that the training time of CZHash is typically faster  than AgNet on these three datasets. Note the runtime of AgNet on Nus-wide is more than 4 days and not reported. This is beacuase AgNet requires a lot of time to separately train the three deep neural networks, which identifies again the superiority of our CZHash jointly optimizing the feature learning, category space learning and hashing quantification processes. Both the convergence curves and runtime analysis can indicate the efficiency of our CZHash.
\vspace{-2pt}
\begin{table}[h!t]
\label{Table8}
\centering
\vspace{-5pt}
\caption{Running times (in seconds) with code length fixed to 16 on three datasets, '-' means the runtime is more than 5 days.}
\begin{tabular}{c|r|r|r}
\hline
  & Wiki  & Mirflickr & Nus-wide \\\hline
$\textbf{AgNet}$ &	168.84 &	3784.65 & --\\
 $\textbf{CZHash}$ & 94.34 &	 2197.63 &  16742.35 \\\hline
\end{tabular}\\
\label{Table8}
\end{table}
\vspace{-5pt}
\section{Conclusion}
\label{sec:concl}
In this paper, we proposed a novel  cross-modal hashing framework (CZHash) to  effectively leverage unlabeled and
labeled multi-modality data with different label spaces for cross-modal zero-shot retrieval. CZSHash integrates deep feature learning, category space learning, and hashing code learning processes. It also introduces a composite similarity to leverage labeled and unlabeled data with different label spaces. Experimental results under different scenarios prove the effectiveness of CZHash. In our future work, we will study more general cross-modal zero-shot hashing approaches for more open tasks.

\section{Acknowledgement}
We appreciate the authors who kindly share their codes and experimental settings with us for experiments.

\bibliographystyle{IEEEtran}
\bibliography{SZCMH_Bib}

% that's all folks
\end{document}